\documentclass[twocolumn]{article}

\usepackage[utf8]{inputenc}
\usepackage{fullpage}
\usepackage{authblk}
\usepackage{url}
\usepackage{hyperref}
\usepackage{multirow}%
\usepackage{amsmath,amssymb,amsfonts}%
\usepackage{amsthm}%
\usepackage{mathrsfs}%
\usepackage[title]{appendix}%
\usepackage{xcolor}%
\usepackage{textcomp}%
\usepackage{manyfoot}%
\usepackage{booktabs}%
\usepackage{algorithm}%
\usepackage{algorithmicx}%
\usepackage{algpseudocode}%
\usepackage{listings}%
\usepackage[inline]{enumitem}
\usepackage{todonotes}
\usepackage{natbib}

\usepackage{tikz}
\usetikzlibrary{shapes.geometric, arrows, positioning, fit, calc}
\tikzstyle{block} = [rectangle, draw, text centered, rounded corners, minimum height=1cm]
\tikzstyle{arrow} = [thick,->,>=stealth]

\newcommand{\fstbest}[1]{\textbf{#1}}
\newcommand{\sndbest}[1]{\underline{#1}}

\def\sndefinitions{
\section*{Acknowledgements}

This work was partially funded by CAPES Finance code (001) and CNPq.
Some of the HPC used in this work was funded by Petrobras call TC 5900.0111175.19.9, 2020/00182-5, and FAPERGS call 16/2551-0000348-8 to Lucas Melo Schnorr. Some experiments in this work used the PCAD infrastructure, \url{http://gppd-hpc.inf.ufrgs.br}, at INF/UFRGS. We would like to thank Dylan Alexander Slavin Hillier for his help proofreading this paper.

\section*{Authors' contributions}

PHdCA did experiment and model design, implementation, and execution, and wrote and reviewed the article. ART and LCL, interacted in discussions, wrote, and reviewed the article.
}

\newcommand{\longpapertitle}{Is an Image Also Worth 16x16=256 Superpixels? A Framework for Attentional Image Classification}

\title{\longpapertitle} %% Article title

\author[ufrgs,uom]{{Pedro Henrique} {da Costa Avelar}\thanks{This work was developed entirely at UFRGS.}\textsuperscript{,}}

\author[ufrgs]{{Anderson R.} {Tavares}}

\author[ufrgs]{{Luís C.} {Lamb}}

\affil[ufrgs]{Institute of Informatics, Federal University of Rio Grande do Sul (UFRGS), 91501-970, Porto Alegre, Rio Grande do Sul, Brazil}
\affil[uom]{Division of Informatics, School of Health Sciences, Imaging and Data Science, Faculty of Biology, Medicine and Health, University of Manchester, M13 9GB, Vaughan House, Portsmouth St, Manchester, United Kingdom}

\date{}

\begin{document}

\maketitle

%% Abstract
\begin{abstract}
Superpixel-based image classification has traditionally leveraged graph neural networks (GNNs) for processing irregular image representations. Recent advances in computer vision, driven by Vision Transformers (ViTs), have introduced new paradigms in self-attentional models, surpassing convolutional neural networks (CNNs) in various tasks. However, a synergistic connection between GNNs, superpixels, and transformers remains unexplored. In this work, we propose Superpixel Transformers (SPT), a novel framework that unifies superpixel-based image classification and ViTs. SPT generalizes the Superpixel Image Classification with Graph Attention Networks (SICGAT) model and ViT to support arbitrary superpixel-based chunking strategies, connectivity graphs, and positional encodings. We introduce refinements including a multidimensional sine-cosine positional encoding and an enriched patch data structure that fully incorporates superpixel shape and color information. By testing SPT across datasets such as CIFAR10, FashionMNIST, and Imagenette, with various superpixel generation and graph connectivity strategies, we demonstrate that SPT achieves superior performance compared to previous superpixel-based GNN methods and remains competitive with ViTs. Notably, our approach addresses the limitations of SICGAT, such as information loss during pixel aggregation, and shows how constrained graph connectivity can enhance ViT performance. SPT bridges the gap between superpixel-based and transformer models, opening avenues for cross-domain generalization and future innovations in hybrid attentional frameworks, and showing that an image can also be worth $16\times16$ superpixels.
\end{abstract}

\section{Introduction}\label{sec1}

Although superpixel-like segmentation algorithms have been used together with Neural Networks for processing images since 2005 \cite{bianchini_recursive_2005}, it was only in 2017 that modern iterations of Graph Neural Networks started using superpixel algorithms to generate image graphs \cite{monti_geometric_2017}. This scheme has been worked on and improved throughout many iterations throughout the years. In particular, using a Graph Neural Network kernel with an attentional mechanism \cite{avelar_multitask_2019} greatly improved performance from previous works even when using a more restrictive adjacency matrix, which the authors claim makes the problem harder to solve.

Shortly after this development of using attention mechanisms and Superpixels for image processing, the uncontested leadership of Convolutional Neural Networks (CNNs) in Computer Vision was upended by another attention-based approach: Vision Transformers \cite{dosovitskiy_image_2021}. Vision Transformers have since become a staple in Computer Vision and, despite being notoriously data-hungry, they have surpassed CNNs in many applications, including image classification, object detection and semantic image segmentation \cite{steiner_how_2022}.

To the best of our knowledge, no previous work has yet linked Vision Transfomers with GNN- and Superpixel-based image classification. In this paper we will bridge the gap between Superpixel Image Classification and Vision Transformers and put them into a framework of attentional image classification. We attempt to ablate which components are crucial for better performance and generalisability in an attempt to challenge the notion that Graph Neural Networks cannot be deep and unify both fields.

We do so by proposing a framework that generalises both the Superpixel Image Classification with Graph Attention Networks (SICGAT) model proposed in \cite{avelar_superpixel_2020} as well as the Vision Transformer (ViT) model \cite{dosovitskiy_image_2021}, while providing a framework that can work with multiple patch types and allowing for arbitrary graph connectivities. We include tests with both grid-patches like in ViT and also extend SICGAT's superpixels into a patch-like data-structure that fully utilises the information from both the superpixel's shape as well as the colour information of each component superpixel. We test our models with three common graph strategies from the literature \cite{monti_geometric_2017,avelar_superpixel_2020,dosovitskiy_image_2021,rodrigues_graph_2024} and show that Vision Transformers might benefit from a more constrained connectivity graph between its patches. Finally, we propose an extension of the Sine-Cosine positional embedding from \cite{vaswani_attention_2017} to an arbitrary number of dimensions, which we show to be highly beneficial for working in a superpixel environment. We hope that these developments will allow the areas of Superpixel Image Classification to be unified with those of mainstream Computer Vision and that these models spur further investigation on different superpixel formats for future ViT work.

\section{Background}
\label{sec:background}

\subsection{Image superpixels}
\label{sec:superpixels}

Superpixels are groups of pixels in semantically similar areas of an image. A number of techniques exist to generate superpixels from images, such as SLIC \cite{achanta_slic_2012}, SNIC \cite{achanta_superpixels_2017}, SEEDS \cite{van_den_bergh_seeds_2012}, ETPS \cite{yao_real-time_2015}, and SH \cite{wei_superpixel_2018}. Out of these techniques, SLIC is widely thought of as one of the most stable and is still recommended over other state-of-the-art algorithms \cite{stutz_superpixels_2018}.

SLIC is a modified version of the k-means clustering algorithm, which only attempts to assign points to cluster centroids if these points are inside a $2S$-sized square centred in the centroid, where $S = N/k$, with $N$ being the number of pixels in the image and $k$ the desired number of superpixels. It works by initialising $k$ cluster means $C_{j}$ composed of a colour component $c_{C_{j}}$ and a spatial component $s_{C_{j}}$, and then iteratively assigning pixels to clusters, and updating the cluster means based on their component pixels. Distances between cluster means and respective and pixels are calculated as the L2 norm of the two normalised Euclidean distance vectors: one for the colour component and one for the spatial component, normalised by the expected maximum spatial distance (S) and the maximum expected colour distance, which is usually represented by a manually-selected constant $m^{2}$.

While the usual representation for images -- a grid/tensor of pixels -- can already be seen as a regular grid graph (a lattice), superpixels induce a more natural graph representation of an image \cite{monti_geometric_2017,avelar_superpixel_2020}. This is due to the fact that, since superpixels are usually irregular, the connectivity between superpixels requires a more abstract structure than a simple grid or hypercube. Thus, superpixels which have bordering pixels become connected between themselves. Therefore, a superpixel-based representation is more suitable for processing with Graph Neural Networks (Section \ref{sec:gnns}).

\subsection{Graph Neural Networks}
\label{sec:gnns}

Graph Neural Networks (GNNs), pioneered by \cite{scarselli_graph_2009}, have had considerable impact on Machine Learning \cite{battaglia_relational_2018}. This is due, in part, to the Graph Convolutional Network of \cite{kipf_semi-supervised_2017}, which made GNNs accessible in modern deep learning frameworks. This has been followed shortly by demonstrations of the power of GNN to model inductive biases of a problem by carefully crafting graphs that matched the structure of the problem \cite{gilmer_neural_2017,battaglia_relational_2018}. A notable example in this line was the development of NeuroSAT, which modelled a boolean satisfiability (SAT) instance into a bipartite graph between literals and clauses, allowing communication to flow between these iteratively until the neural ``solver'' reaches equilibrium \cite{selsam_learning_2019}.

GNNs can be interpreted through many different lenses, of which the most common are either a generalization of convolutions and convolutional neural networks \cite{monti_geometric_2017,kipf_semi-supervised_2017,fey_splinecnn_2018}, or by interpreting GNNs as what are called ``Message-Passing Neural Networks'' (MPNN) \cite{gilmer_neural_2017,selsam_learning_2019}. In summary, under the MPNN framework, a simple graph neural network will receive a graph $G = \{V,E\}$ composed of a set of vertices $v \in V$ and edges $e \in E$, with edges connecting a source $s \in V$ and a target node $t \in V$ $e = (s,t)$. The neural network will first embed the attributes of each of the nodes through a usually learned function so that we have $x_{v}^{0} = f(v)$ and then propagate the information between the nodes using the adjacencies defined through the edges. Under the MPNN framework, this is seen as each of the nodes $u$ being a neural computer $g$ with state $x_{u}$ which then exchanges messages $m_{u}^{t} = g^{t}(x_{u}^{t})$ at a discrete time step $t$ (which can also be interpreted as a layer). These messages then are aggregated through some, usually order-invariant, function $a$ on the node's neighbourhood $N(u)$ to form the basis of the next time steps' state $x_{u}^{t+1} = a(\{m_{v}^{t+1}, v \in N(u)\})$

\subsection{Self-attention in Sequences, Images and Graphs}

Self-attention establishes connections between different positions of a sequence or set to construct representations, and one of the most successful earlier applications was to provide a learned mapping from sentences between different languages during machine translation \cite{bahdanau_neural_2015}. This approach was then successfully applied in many different tasks, including providing approximations to NP-Complete problems such the Travelling Salesperson Problem \cite{vinyals_pointer_2015}. However, most of these earlier methods still used other architectural patterns, mostly RNNs, whose input was mediated through attention.

Soon after the development of self-attentional mechanisms to mediate the sequences generated through RNNs, the development of the Transformer model \cite{vaswani_attention_2017} showcased the power behind attentional mechanisms. The Transfomer exclusively uses self-attention to mediate the exchange of information between elements of a sequence, which mitigated common RNN issues, such as challenges with long-time dependencies \cite{schuster_bidirectional_1997,graves_bidirectional_2005} and the inherent lack of parallelism \cite{vaswani_attention_2017}.
This new architectural pattern has then been widely adopted and has had applications ranging from Natural Language Processing \cite{openai_gpt-4_2023}, to Computational Chemistry \cite{wu_molformer_2023,honda_smiles_2019,wang_smiles-bert_2019}, Computational Biology \cite{chen_transformer_2023,cui_scgpt_2023} and, of course, has also recently been widely applied in image processing through the development of the Vision Transformer \cite{dosovitskiy_image_2021}.

The self-attention mechanism in text-based networks are often mediated both through a positional embedding, as well as an attention mask which limits which values in the sequence an element can attend to. The original Transformer, due to its language model purpose, limits the attention mechanism so that elements can only attend to those before it, and uses a Sine-Cosine expansion depending on the position of the element in the sequence to help the attention heads to identify the distance between two attended elements. Another option which also became widespread was that of using Bidirectional Encoder Representations from Transformers (BERT) \cite{devlin_bert_2019}, allowing all elements to attend to all others, and using a learned positional embedding instead, which was deemed more appropriate for the vision context and used in the Vision Transformer paper \cite{dosovitskiy_image_2021}.

\subsubsection{Graph Attention Networks}

Graph Attention Networks (GATs) \cite{velickovic_graph_2018} were originally developed in 2017\footnote{Arxiv version released in 2017 \url{https://arxiv.org/abs/1710.10903}} after a surge of interest in Graph-based neural networks \cite{battaglia_relational_2018} due to in part to the Graph Convolutional Network model \cite{kipf_semi-supervised_2017} making graph neural networks accessible in modern deep learning frameworks. Graph attention networks function in a very similar way to other GNN architectures, except that they mediate the message-passing between neighbours through a self-attention mechanism, allowing the network to learn how to prioritise information flow depending on the each node's information content. This architecture was further refined to allow for the attentional mechanism to be utilised to a higher potential \cite{brody_how_2022}.

Is is possible, however, to draw a relationship between GATs and Transformers. In fact, if we assume that a Transformer's attention mask can be treated as the adjacency matrix in a GAT, it is clear that both architectures work in a similar way, with a layer transforming each input/node, and then having self-attention mediating the information between non-masked inputs/neighbours. GATs differs from Transformers almost only through the lack of specific layers in the update block, such as residual/skip connections and layer/batch normalisation of the input. This similarity has been exploited to develop graph-specific transformer architectures through adopting a graph-specific positional embedding using Laplacian eigenvectors \cite{dwivedi_generalization_2021}.

\section{Related Work}

While an image is usually represented as a grid of pixels, as mentioned above, if one were to work using superpixels instead, a graph representation becomes required due to the irregular connectivity pattens between superpixels, and has been widely used together with Graph Neural Networks for Superpixel Image Classification \cite{monti_geometric_2017,fey_splinecnn_2018,avelar_superpixel_2020,rodrigues_graph_2024}. However, to allow for such general connectivity patterns, GNN-based methods often aggregate a node’s neighbourhood using the same weight for every neighbour. This implies that simple GNNs would have a lower representational power than convolutional neural networks, since CNNs allow for a different weight to be used for each value in the ``neighbourhood'' (the convolution window). In fact, a pixel-grid-based GNN was tested in \cite{monti_geometric_2017} and had a worse performance than a CNN.

The model presented in Superpixel Image Classification with Graph Attention Networks, here onwards referred to as SICGAT, \cite{avelar_superpixel_2020} expanded on previous work to allow for a GNN that enables a pseudo-convolutional representational power. This was achieved through using multiple attention heads to mediate the information transfer between nodes in a graph, which allows different ``soft''-weights to be learned by the network when aggregating between neighbourhoods. Surprisingly, however, the performance of these models reached its limit with a much smaller amount of attention heads than the medium connectivity in the superpixel graph they built (which is a Region Adjacency Graph). In this paper we will provide a generalisation of their method and show that our model has better results even than their ``hard'' baseline of using a VGG architecture that only has access to a superpixel-segmented image where each superpixel's pixels have their RGB values set to their common average. Our model also overcomes the information loss they show in Figure~\ref{fig:superpixel_information_loss}.

Recently there has also been a comparative study on the best graph construction method for GCNs \cite{rodrigues_graph_2024}, which showed an improved performance on \cite{avelar_superpixel_2020}'s work even when using a GNN with a lower representational power \cite{xu_how_2019}. They provide a thorough study showing an ablation of various parameters used to build a GNN for SIC and argue that performance was capped at very shallow layer-depths. In this paper we will also perform a thorough exploration of an attention-based GNN model which, thus, has a higher representational power than GCNs and is at least as powerful as a Weissfeiler-Lehman test \cite{xu_how_2019}. Through our ablation, we show that, although more layers did not always cause better performance, the layer depth of our best-performing model often reached our cap for the number of layers (12), which allows for networks four times as deep as the ones used in \citet{avelar_superpixel_2020} and \citet{rodrigues_graph_2024}.

\section{Superpixel Transformers: A General Framework for Attentional Learning in Images}

In this section we provide the definition of our model that generalises the SICGAT model the ViT architecture. We first start by providing a step-by-step description of the differences between SICGAT and the ViT architectures, and then we follow with a description of a concrete formalisation of our model.

\subsection{Key differences between Superpixel Image Classification and Vision Transformers}

\begin{figure*}
    \centering
    \begin{tabular}{cc}
    \includegraphics[width=0.45\linewidth]{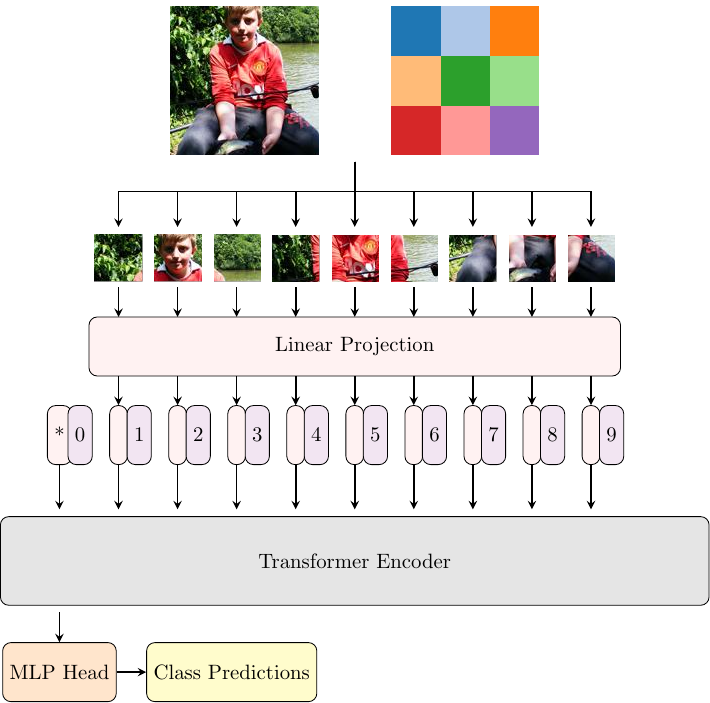}
        %\resizebox{0.45\linewidth}{!}{%
        %    \input{diagram-vit.tex}
        %}
            &
    \includegraphics[width=0.45\linewidth]{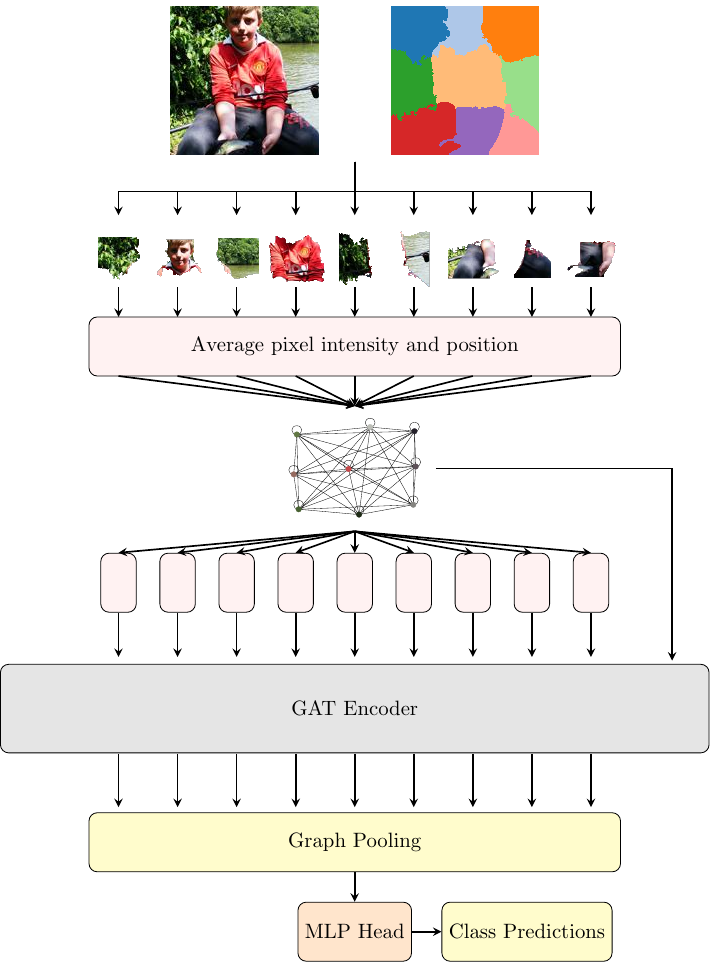}
        %\resizebox{0.45\linewidth}{!}{%
        %    \input{diagram-sicgat.tex}
        %} 
    \end{tabular}
    \caption{The ViT architecture \cite{dosovitskiy_image_2021} (left) and the SICGAT architecture \cite{avelar_superpixel_2020} (right). Note that, apart from the main self-attention encoding layer, the adjacency graph masking and the fact that SICGAT compresses all input through averaging the channel information of the superpixel instead of having access to all pixels, both architectures are remarkably similar.}
    \label{fig:SICGAT_vs_ViT_arch}
\end{figure*}

\begin{figure}
    \centering
    \begin{tabular}{cc}
    \includegraphics[width=0.33\linewidth]{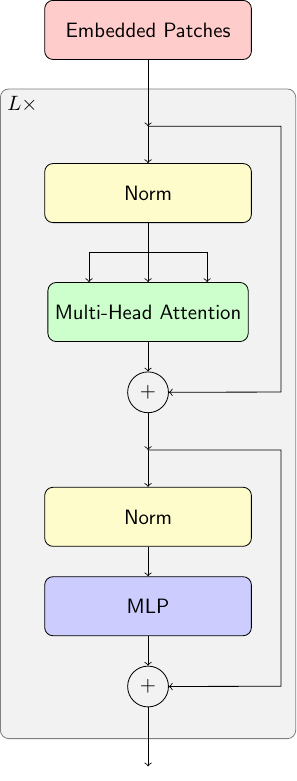}
        %\resizebox{0.15\linewidth}{!}
        %{%
        %    \input{diagram-vit-enc.tex}
        %} 
            &
    \includegraphics[width=0.33\linewidth]{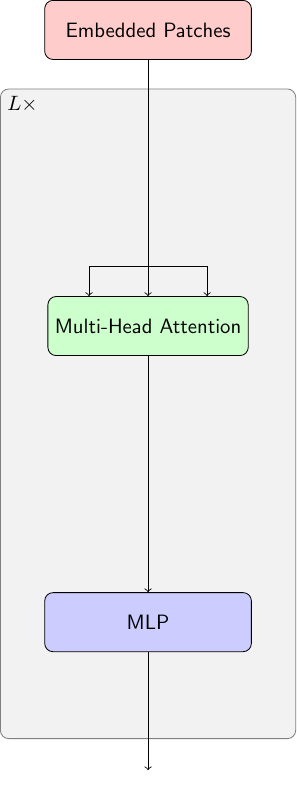}
        %\resizebox{0.15\linewidth}{!}
        %{%
        %    \input{diagram-sicgat-enc.tex}
        %} 
    \end{tabular}
    \caption{The Transformer Encoder \cite{vaswani_attention_2017} used on the ViT model \cite{dosovitskiy_image_2021} (left) and an equivalent view of a GAT encoder \cite{velickovic_graph_2018} used on the SICGAT model \cite{avelar_superpixel_2020} (right). Using an attention mask on a transformer encoder would then be equivalent to using an adjacency amtrix on a GAT encoder, with the only exception being the inner normalisation layers and residual connections. 
    }
    \label{fig:SICGAT_vs_ViT_enc}
\end{figure}

Figure~\ref{fig:SICGAT_vs_ViT_arch} shows a comparison of the SICGAT and ViT architectures. From this, we identify the main differences between these two approaches: (1) Chunking strategy, (2) pixel information aggregation, (3) positional embeddings, (4) architectural differences between a GAT and a Transformer block, (5) differences in chunk connectivity, and (6) pooling vs virtual classification node. We discuss in depth each of these differences next.

\textbf{Chunking Strategy} The first and most obvious difference between the two models is the fact that while SICGAT uses a superpixel algorithm to chunk the image into superpixels, ViT chunks images into equally-sized rectangles. However, this scheme had also previously been used in other superpixel image classification algorithms, such as in MoNET \cite{monti_geometric_2017}, and the approach followed by SICGAT could easily be adapted to also use equally sized chunks. The main reason, however, why SICGAT performs better when chunking the image into superpixels instead of regular patches is due to the pixel information aggregation scheme.

\begin{figure*}
    \centering
    \includegraphics[width=.75\textwidth]{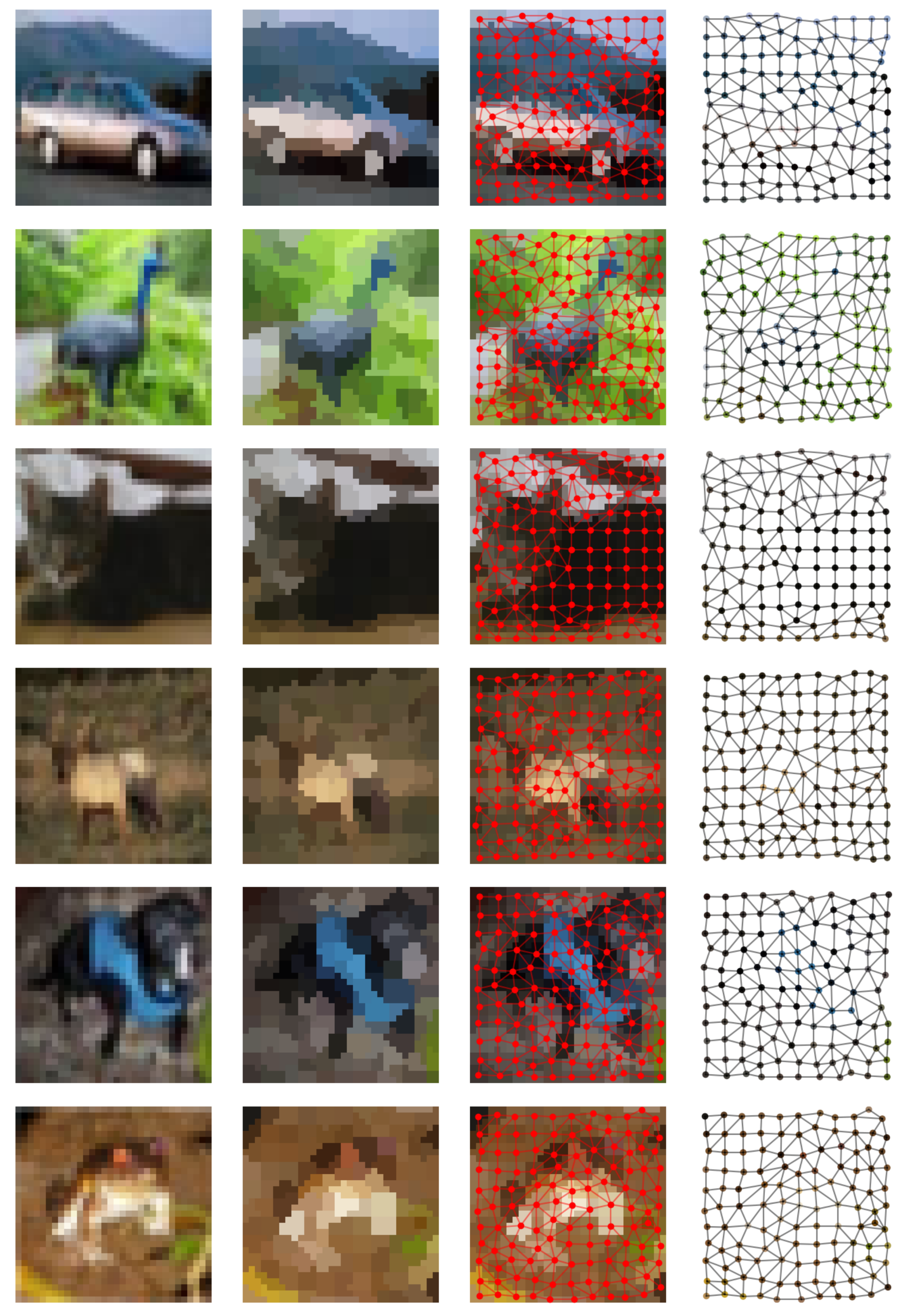}
    \caption{The information loss that happens when converting an image to a superpixel representation and then to its equivalent graph representation.}
    \label{fig:superpixel_information_loss}
\end{figure*}

\textbf{Pixel Information Aggregation} Most superpixel image classification methods aggregated the information from all of a chunk's pixels into a single value, such as taking the average value of pixel intensities around the pixel. This was also followed by SICGAT, and it was shown to cause heavy information loss, as can be seen in Figure~\ref{fig:superpixel_information_loss}. ViTs mitigate this by making a linear projection of the flattened pixel inputs for each chunk, which allows the network to learn to use the information from every pixel in a chunk, mitigating the information loss present in SICGAT and other methods such as MoNet.

\textbf{Positional Embeddings} Another difference between these approaches is how positional information is treated. Earlier superpixel image classification models such as MoNet used positional information only as a weighting factor when performing neighbourhood aggregation, simply using this information as an inductive bias for the network. SICGAT extended this by appending each node's positional information to the node's features, treating the positional information in the same way as its pixel information. ViTs follow similarly but with two major differences: first, instead of appending the positional information to the pixel information, ViTs follow the previous standard used by Transformers and sums the positional information with the pixel information; lastly, due to the fact that ViTs uses a fixed number of equidistant patches, it encodes the positional information directly through a patch's index instead of as a transformation of its xy coordinates.

\textbf{Architectural differences} A GAT layer and a Transformer layer have significant differences in their architectural components, with GATs being, in a sense, a simpler form of a Transformer Layer, which can be seen in Figure~\ref{fig:SICGAT_vs_ViT_enc}. Firstly, The GAT layer using in SICGAT only performs multi-head attention and immediately follows it with a linear/MLP block to update a node's information. The transformer layer instead first norms information before either the multi-head attention or MLP blocks, and then uses a residual approach to update the information instead of directly editing it, which has been shown to be an easier problem to solve and, thus, more efficient parameter-wise.

\textbf{Chunk Connectivity} Another difference is how connectivity is handled. Many of the earlier works in Superpixel Image Classification followed the same connectivity scheme as done by MoNet: a KNN graph based on chunk position. SICGAT tested their architecture only on a more restricted version of the problem, using instead a Region Adjacency Graph (RAG), arguing that it was both a harder problem as well as being closed conceptually to a convolution block, since only chunks that were physically connected would be also connected in this scheme. ViTs eschew this altogether and treat the image as a fully connected graph, leaving for the attentional mechanism to route the information flow between chunks.

\textbf{Pooling vs virtual classification node} Finally, the last major difference between these two models is that, while SICGAT average-pools the information in all nodes before feeding it into a classification head, ViTs again followed the traditional Transformer approach of addeding a virtual classification node, which is then passed to the classification head. This would be equivalent to adding a virtual graph node, which is also another commonly used way to perform graph classification in Graph Neural Networks. In this paper we will follow the original ViT paper and work mainly with the concept of the classification, however it is worthy of note that more recent papers claim better performance by classifying and pooling over every single patch instead \cite{beyer_better_2022}.

With these differences in mind, we can then build a generalisation that allows for both models to be represented under the framework, which we'll describe in the following sections.

\subsection{The Superpixel Transformer Model}

\begin{figure*}
    \centering
    \includegraphics[width=0.90\linewidth]{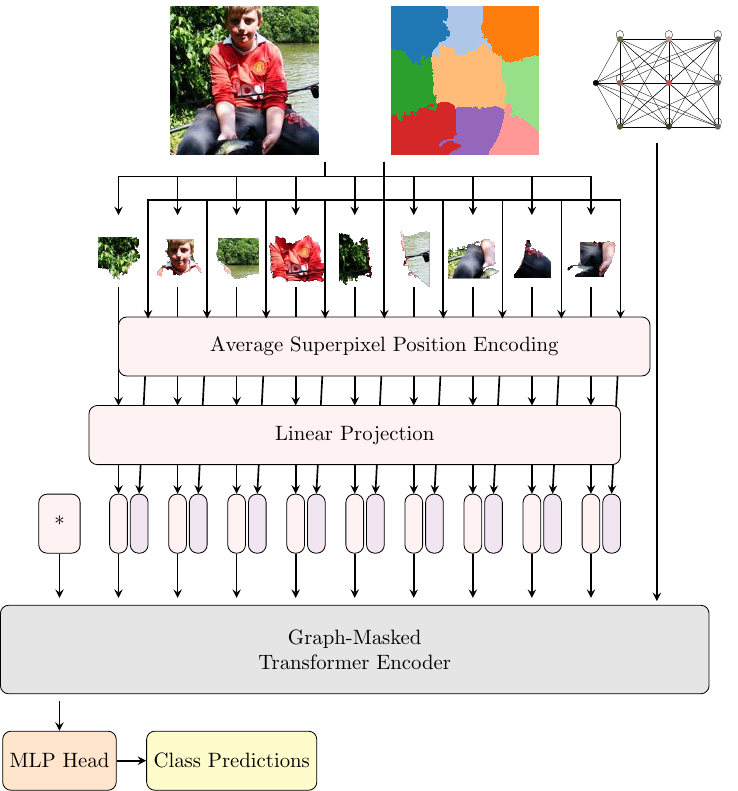}
    %\resizebox{0.90\linewidth}{!}{%
    %    \input{diagram-spt.tex}
    %}
    \caption{Our proposed Superpixel Transformer (SPT) architecture. Differently from SICGAT \cite{avelar_superpixel_2020} we use the full superpixel information through the patch data structure built in Algorithm~\ref{alg:getsuperpixelpatches} and use a variety of superpixel positional encodings to build the input for our encoder network instead of encoding the position together with the patch informaiton in the first GAT layer. Different from ViT \cite{dosovitskiy_image_2021} we allow for arbitrary superpixel algorithms and also use our model with a variety of adjacency graphs which constrain and help information flow better through the model. Also, due to the superpixel patches possibly having different shapes and positions, we extend and allow for encodings that use the xy coordinates of the superpixel/patch.}
    \label{fig:SPT}
\end{figure*}

We define here our framework called Superpixel Transformers (SPT) (Figure~\ref{fig:SPT}), which are an attempt at generalising (mostly attentional) SIC models and bridging the gap between these models and Vision Transformers, which see use in mainstream computer vision. Our framework is composed of 6 steps divided into preprocessing and on how to use the input for the network architecture. In the preprocessing stage, given an image $I$, we use a chunking strategy $f_{S}$ to divide the image into its chunks/superpixels $S$ that contain information about both the superpixel's content $S_c$ and its position $S_p$ on the image, then we use a connectivity strategy $f_{G}$ to build a connectivity graph $G$ between the superpixels. Then, the superpixels $S$ and input graph $G$ are passed first encoded through a positional encoding strategy $P$ and a content encoding strategy $C$, to be passed through a self-attention module (e.g. Transformer Encoder) that will finally be used to produce a classification. We'll discuss the particular choices for each step below.

In the preprocessing part we choose two specific chunking strategies: A grid-based chunking (similar to what was done in \cite{dosovitskiy_image_2021}) that we define precisely in Algorithm~\ref{alg:squaregrid} and which returns a $\mathbb{Z}^{h \times w}$ integer mask showing which superpixel each pixel belongs to, and SLIC0 \cite{achanta_slic_2010,achanta_slic_2012} (here onwards also referred to as SLIC), which provides a similar output but follows a spatiocontent kmeans variation to divide pixels into superpixels. Specifically, we used the implementation in \texttt{scikit-image} version 0.24. After obtaining the mask from a superpixel strategy, such as one of the two above, we use the segment mask to construct a neighbourhood graph $G$ and use both the image and the segment mask into an algorithm that builds out superpixel patch data structure. The data structure we use is composed of three major components: An $\mathbb{R}^{k \times 2}$ tensor containing the xy positions of each superpixel, an $\mathbb{R}^{k \times 3 \times h_{chunk} \times w_{chunk}}$ tensor that contains the superpixel's RGB content, and an $\{0,1\}^{k \times 1 \times h_{chunk} \times w_{chunk}}$ mask that contains the superpixel's shape content. We originally used a $h_{chunk}=w_{chunk}=k_{chunk}\times S$ with $k_{chunk}=2$ since that is the search limit on the SLIC algorithm \cite{achanta_slic_2010}, but the SLIC implementation we used allowed superpixels with dimensions larger than $2S$, therefore we set $k_{chunk}=3$. To allow for batching we provide a fixed-size superpixel chunk tensor, and thus we also provide a $\{0,1\}^{k}$ mask to inform the transformer which chunks contain information. The full procedure to build this data structure can be seen in Algorithm~\ref{alg:getsuperpixelpatches}. Note that while we define our model here for 2-dimensional data with 3 channels, it can easily be adapted to more dimensions and a different number of channels. For simplicity's sake we transform any image with only a single channel to a 3-channel grayscale image.

\begin{algorithm*}
\caption{Get Superpixel Patches}
\label{alg:getsuperpixelpatches}
\begin{algorithmic}[1]
\Require $\text{image} \in \mathbb{R}^{C \times h \times w}$, $\text{segments} \in \mathbb{Z}^{h \times w}$, $\text{max\_patches} \in \mathbb{Z}$, $\text{patch\_shape} \in \mathbb{Z}^2$
\Require $\text{search\_patch\_size} \in \mathbb{Z}$, $\text{background\_fill}$ strategy
\Ensure $\text{spixpresent}, \text{spixcenters}, \text{spixpatches}, \text{spixpatchpresent}$

\State Initialize $\text{spixpresent}$, $\text{spixcenters}$, $\text{spixpatches}$, and $\text{spixpatchpresent}$ if not provided
\State Fill $\text{spixpatches}$ according to a $\text{background\_fill}$ strategy \Comment{Default: all zeros}
\State Mark segments with data: $\text{spixpresent}[:\text{segments.max}() + 1] \gets 1$
\For{$i \gets 0$ to $\text{segments.max}()$}
    \State $\text{spixpos} \gets \text{positions where} \ \text{segments} = i$
    \State $\text{spixmins}, \text{spixmaxs} \gets \text{min}(\text{spixpos}), \text{max}(\text{spixpos})$
    \State $\text{spixcenters}[i] \gets \text{spixmins} + \frac{\text{spixmaxs} - \text{spixmins}}{2}$
    \State $y_{0}, x_{0} \gets \lfloor(\text{spixcenters}[i] - \frac{\text{search\_patch\_size}}{2}) \rfloor$
    \State $y_{1}, x_{1} \gets \lfloor(\text{spixcenters}[i] + \frac{\text{search\_patch\_size}}{2}) \rfloor$
    \State $\text{spixpos} \gets \text{positions within box} \ [y_{0}, y_{1}, x_{0}, x_{1}]$
    \State Update $\text{spixpatches}$ by copying colour information from the $i$-th superpixel's component pixels
    \State Update $\text{spixpatchpresent}$ as a binary mask showing which pixels in $\text{spixpatches}$ are contained in $i$-th the superpixel
\EndFor

\Return $\text{spixpresent}, \text{spixcenters}, \text{spixpatches}, \text{spixpatchpresent}$
\end{algorithmic}
\end{algorithm*}

For the graph-generation strategies we tested three strategies: 
\begin{enumerate*}
    \item Using a fully connected graph \texttt{fcg}, similar to what a Vision Transfomer uses \cite{dosovitskiy_image_2021};
    \item Using a graph build through a positional k-nearest neighbours \texttt{knn} \cite{monti_geometric_2017};
    \item And using a Region-Adjacency graph \texttt{rag}, as defined in \cite{avelar_superpixel_2020}.
\end{enumerate*}
These strategies encompass most of the strategies used in previous models.

After producing our data structures it is then fed to our end-to-end model, which encode the chunk's with single fully connected layer as is done with the original ViT \cite{dosovitskiy_image_2021}. Although our model can also be trained using convolutional stems \cite{xiao_early_2021}, we do not do so for this paper. We test three different strategies for encoding a node's position:
\begin{enumerate*}
    \item The original learned positional embedding used in the ViT \cite{dosovitskiy_image_2021} and proposed by \cite{devlin_bert_2019};
    \item a simple Linear layer that maps the xy coordinates to the size of the embedding; and
    \item an adaptation of the Sine-Cosine positional embedding originally used in \cite{vaswani_attention_2017}, but expanded to work with multiple dimensions, the algorithm for which can be see in \ref{alg:xysincos}.
\end{enumerate*}
We then apply a transformer encoder network using the adjacency graph to define the attention mask, as well as an extra classification token added which connects to all other tokens. We chose the transformer encoder architecture instead of a GAT encoder due to both more mature Transformer implementations and the fact that the Transformer Encoder includes normalisation layers which are known to help stabilise training. Finally, we extract the classification token to pass on to a classification MLP.

\section{Environment}

We train all of our models from scratch, taking most of the hyperparameters from \cite{steiner_how_2022}, with different values marked by *: 
\begin{itemize}
    \item Learning Rate: 1e-3
    \item Learning Rate Scheduler*: We Reduce the LR on a plateau by a factor of  with a patience of 20\% the number of epochs
    \item Dropout = 0.1
    \item Global Gradient Clipping = 1.0 for FashionMNIST and Imagenette
    \item Weight Decay: 1e-4
    \item Optimiser: Adam \cite{kingma_adam_2015}, using $\beta1 = 0.9$ and $\beta2 = 0.999$
    \item Due to instabilities on FashionMNIST, we do not backpropagate any batches that cause invalid gradients.
\end{itemize}

Due to compute constraints, we evaluate our models only on 4 datasets: FashionMNIST \cite{xiao_fashion-mnist_2017}, CIFAR10 \cite{krizhevsky_learning_2009}, Imagenette \cite{howard_fastaiimagenette_2024}, and Resisc45 \cite{cheng_remote_2017}, which are described in Table~\ref{tab:dsets}. We have to limit the batch size, setting it as 512 for the MNIST and CIFAR10 datasets, 64 for the TinyImagenet and 32 for the Resisc dataset. We train Ti-sized models on all datasets except on Resisc45, where we train a B-sized model. We use k=15 for all KNN graphs. Whenever a superpixel would violate a patch's bounds, we clip it to fit the patch by keeping the centermost pixels, and we used patch of size $3\times S$.

\begin{table*}[]
    \scriptsize
    \centering
    \begin{tabular}{lrrrrcrr}
    \toprule
        Dataset & Training & Validation & Test & Total & C,W,H & \#Classes & $S$ \\
    \midrule
        FashionMNIST \cite{xiao_fashion-mnist_2017} & *54,000 & *6,000 & 10,000 & 70,000 & $1,28,28$ & 10 & 4 \\
        CIFAR10\cite{krizhevsky_learning_2009} & *47,500 & *2,500 & 10,000 & 60,000 & 3,32,32 & 10 & 4\\
        Imagenette \cite{deng_imagenet_2009,howard_fastaiimagenette_2024} & *8,996 & *473 & 3,925 & 13,394 & $3,160,160$ & 10 & 10 \\
        Resisc45 \cite{cheng_remote_2017} & 28350 & 3150 & $\dagger$  & 31500 & $3,256,256$ & 45 & 10 \\
    \bottomrule
    \end{tabular}
    \caption{The description of the datasets used. C,W,H stand for the number of Channels, Width, and Height of the images. * Original development split was split randomly into train and validation, $\dagger$ No test set provided, results are shown for validation}
    \label{tab:dsets}
\end{table*}

\section{Results}

\begin{figure*}
    \centering
    \includegraphics[width=0.45\linewidth]{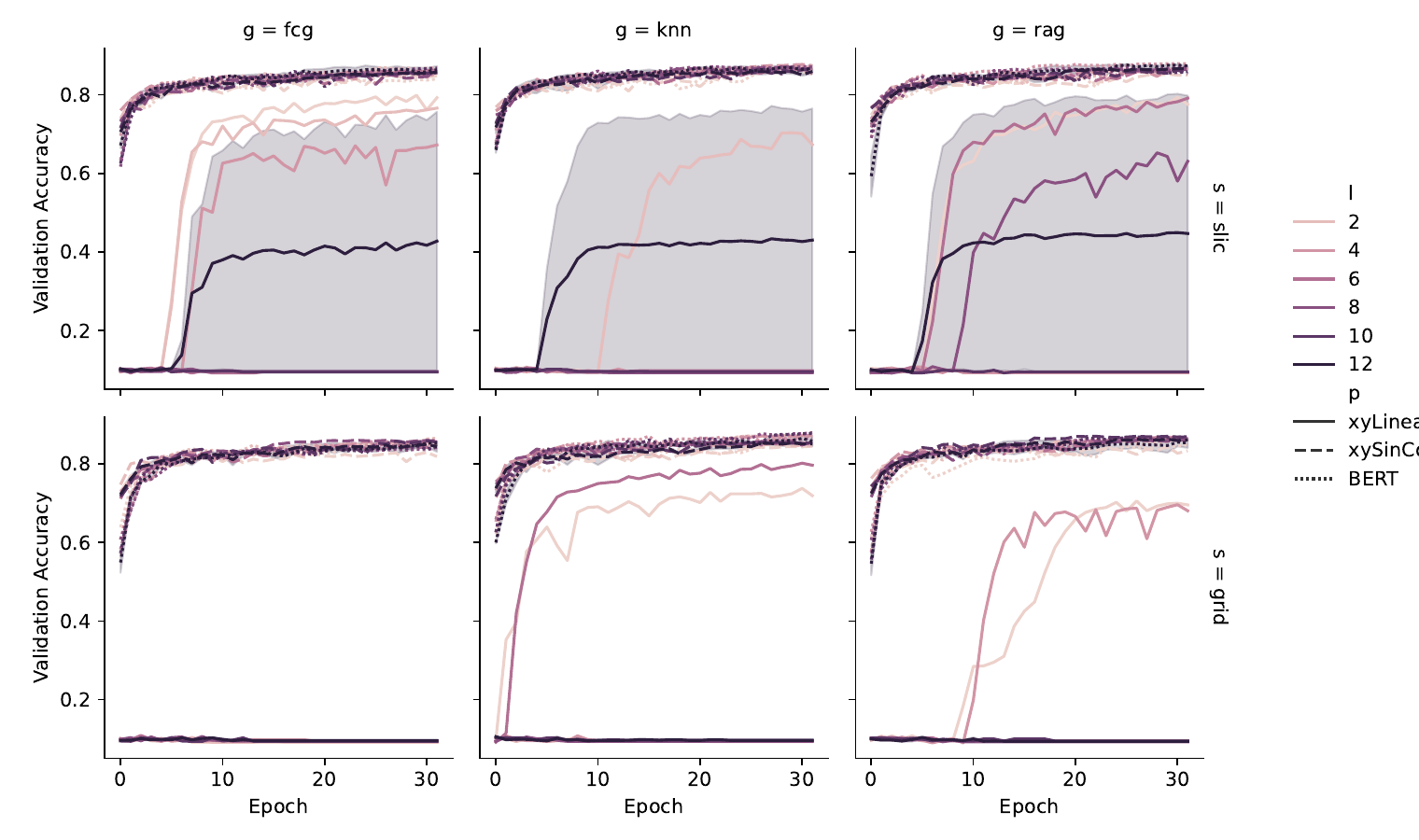}
    \includegraphics[width=0.45\linewidth]{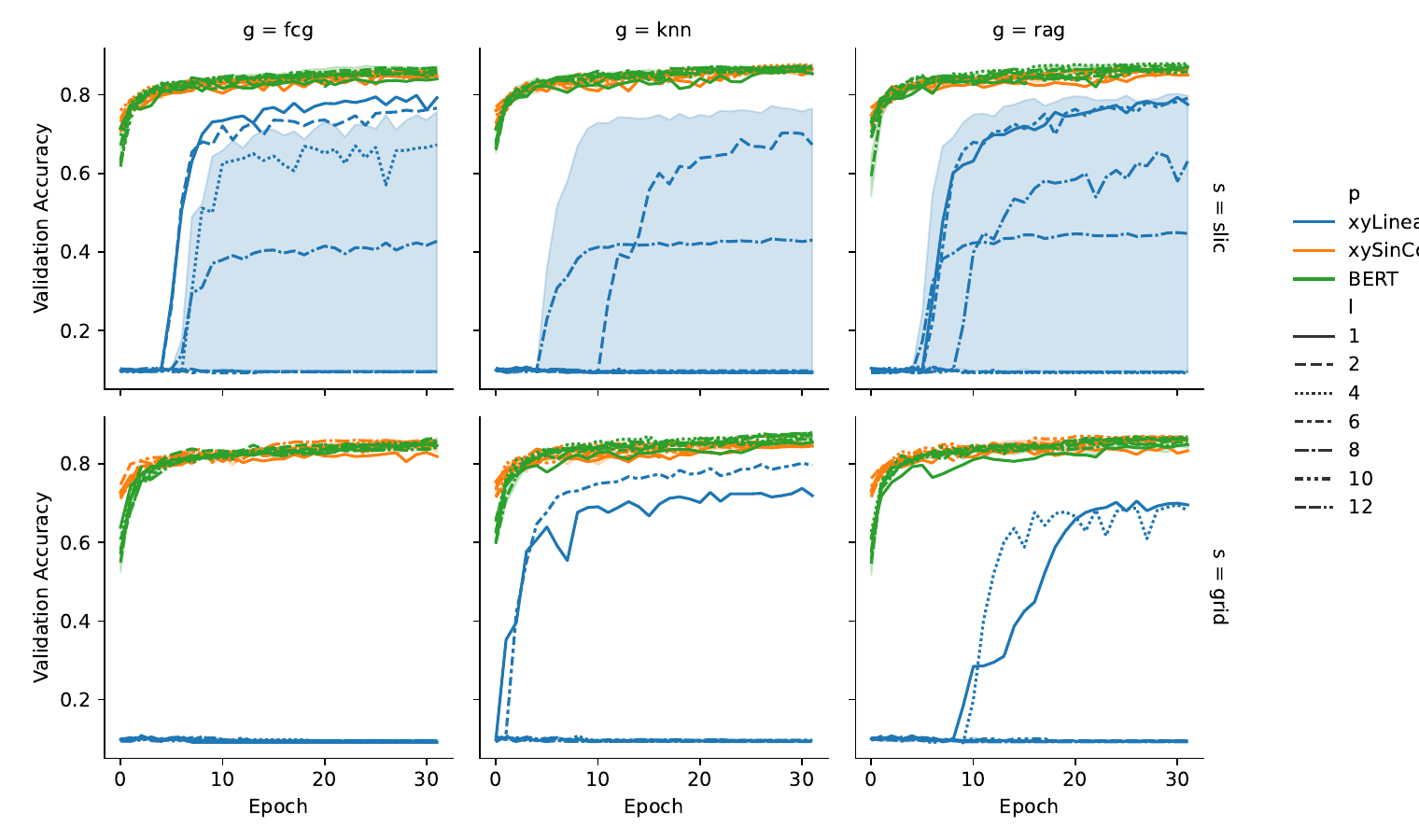}
    \caption{Plots highlighting the Validation Accuracy throughout the epochs for all the configurations we trained on the FashionMNIST dataset. One can see from graph on the left that the number of layers seemed to have little to no impact for the FashionMNIST dataset, except for the Linear positional embedding based on the xy coordinates. From the graph on the right it's possible to see that the Linear embedding caused an extremely unstable training and sometimes even failed to reach similar accuracy levels.}
    \label{fig:fashionmnist-ablation}
\end{figure*}

We trained Ti-sized models \cite{steiner_how_2022} using all combinations of graph connectivity strategies, patch building strategies, and superpixel building strategy, for 7 possible layer sizes: 1, 2, 4, 6, 8, 10, and 12. The layer size which provided the best validation accuracy out of the 7 test is shown for each of the combinations in Table~\ref{tab:ablation} and in \ref{fig:fashionmnist-ablation}. The best result of the SLIC-based model from this table is shown in \ref{tab:results} as our model, even though we had a better performing model using the grid superpixel generation strategy. The ViT model (Grid-BERT-\texttt{fcg} combination) achieved a slightly higher but comparable validation performance to our best SLIC model on the CIFAR10 dataset and a slightly lower but comparable performance in the FashionMNIST dataset.

Our proposed graph connectivity strategies also boosted the validation performance of the Grid-based superpixel models when compared to the fully connected graphs, and all of our trained models (including the ViT) outperformed the Graph Superpixel Image Classification baselines in all datasets. We trained an SPT-Ti (12 layers, Sine-Cosine, \texttt{fcg}) and a ViT-Ti (12 layers, BERT, \texttt{fcg}) model on the Imagenette dataset to provide a result for which the image was not too small, and both models were comparable and, while no previous SIC work has provided results on higher-resolution images, we believe that due the information loss shown in \cite{avelar_superpixel_2020} would make it very hard for these models to perform anywhere close to our models.

\begin{table*}
    \centering
    \begin{tabular}{lrrrrrr}
    \toprule
        \multirow{2}{*}{Model} & \multicolumn{2}{c}{FashionMNIST} & \multicolumn{2}{c}{CIFAR10} & \multicolumn{2}{c}{Imagenette} \\
         & Valid & Test & Valid & Test & Valid & Test \\
    \midrule
        VGG-16 \cite{tang_image_2022} & - & 92.0 & - & $77.2$ & 83.3 & 83.5 \\
        VGG-16 SLIC \cite{avelar_superpixel_2020} & - & - & - & $62.9$ & - & - \\
    \midrule
    \midrule
        SICGAT \cite{avelar_superpixel_2020} & - & $83.1$ & $53.4$ & $45.9$ & - & - \\
        gSLICr-GCN (l=12) \cite{tang_image_2022} & - & $81.9$ & - & $59.5$ & - & - \\
        bestGCN \cite{rodrigues_graph_2024} & - & $84.2$ & - & $58.5$ & - & - \\
    \midrule
        ViT \cite{dosovitskiy_image_2021} (our results) & 86.5 & \sndbest{85.6} & 70.6 & \fstbest{69.0} & - & \fstbest{67.9} \\
        SPT-slic (ours) & $87.9$ & \fstbest{87.3} & $67.0$ & \sndbest{63.3} & - & \sndbest{67.7} \\
    \bottomrule
    \end{tabular}
    \caption{Results for the three main datasets we tested our models in. Due to compute limitations we did not re-train models and instead use other author's claimed results. For our models we limit ourselves to using only SPT models using SLIC superpixels and we use the test results of whichever model provided the best valid accuracy in our ablation study. For the ViT model we only consider its original implementation's hyperparameters: Fully-connected graph with a BERT embedding. For the Imagenette result we use fully connected graphs with the Sine-Cosine positional embedding. For the VGG-16 architecture we use results reported in \cite{tang_image_2022,mandal_vgg16_2023,kasen_vgg_2024}.
    }
    \label{tab:results}
\end{table*}

\begin{table*}
    %\footnotesize
    \centering
    \begin{tabular}{lllrrrrr}
    \toprule
        \multirow{2}{*}{S} & \multirow{2}{*}{P} & \multirow{2}{*}{G} & \multicolumn{2}{c}{Valid Acc (\%)} & \multicolumn{2}{c}{Epochs} & \multirow{2}{*}{Layers} \\
        & & & Best & Last & Best & Last & \\
    \midrule
        \multicolumn{8}{c}{FashionMNIST} \\
    \midrule
        \multirow{9}{*}{SLIC} & \multirow{3}{*}{Linear}
            & \texttt{fcg} & 80.0 & 79.2 & 30 & 32 & 1 \\
          & & \texttt{knn} & 77.2 & 76.5 & 28 & 32 & 12 \\
          & & \texttt{rag} & 80.2 & 79.8 & 31 & 32 & 12 \\
        & \multirow{3}{*}{SinCos}
            & \texttt{fcg} & 86.3 & 86.3 & 32 & 32 & 6 \\
          & & \texttt{knn} & 87.7 & 87.4 & 31 & 32 & 4 \\
          & & \texttt{rag} & 87.0 & 87.0 & 32 & 32 & 12 \\
        & \multirow{3}{*}{BERT}
            & \texttt{fcg} & 87.4 & 87.3 & 25 & 32 & 12 \\
          & & \texttt{knn} & 87.5 & 87.5 & 32 & 32 & 12 \\
          & & \texttt{rag} & *\fstbest{87.9} & 87.8 & 31 & 32 & 4 \\
          \\
        \multirow{9}{*}{Grid} & \multirow{3}{*}{Linear}
            & \texttt{fcg} & 10.9 & 9.5 & 3 & 32 & 6 \\
          & & \texttt{knn} & 80.2 & 79.6 & 31 & 32 & 6 \\
          & & \texttt{rag} & 70.5 & 69.5 & 27 & 32 & 1 \\
        & \multirow{3}{*}{SinCos}
            & \texttt{fcg} & 86.0 & 85.9 & 24 & 32 & 8 \\
          & & \texttt{knn} & 87.2 & 87.2 & 32 & 32 & 4 \\
          & & \texttt{rag} & 87.0 & 86.8 & 23 & 32 & 10 \\
        & \multirow{3}{*}{BERT}
            & \texttt{fcg} & $\dagger86.5$ & 86.5 & 32 & 32 & 2 \\
          & & \texttt{knn} & *\sndbest{87.9} & 87.9 & 32 & 32 & 8 \\
          & & \texttt{rag} & 87.2 & 87.2 & 31 & 32 & 2 \\
    \midrule
         \multicolumn{8}{c}{CIFAR10} \\
    \midrule
        %\multirow{9}{*}{SLIC}
        %& \multirow{3}{*}{Linear} 
        %  & \texttt{fcg} & - & - & - & - & - \\
        %& & \texttt{knn} & - & - & - & - & - \\
        %& & \texttt{rag} & - & - & - & - & - \\
        \multirow{6}{*}{SLIC}
        & \multirow{3}{*}{SinCos}
          & \texttt{fcg} & *67.0 & 64.4 & 221 & 279 & 10 \\
        & & \texttt{knn} & 67.0 & 65.7 & 185 & 187 & 2 \\
        & & \texttt{rag} & 66.4 & 65.0 & 279 & 308 & 4 \\
        & \multirow{3}{*}{BERT}
          & \texttt{fcg} & 60.8 & 59.9 & 162 & 271 & 6 \\
        & & \texttt{knn} & 61.9 & 59.6 & 172 & 198 & 10 \\
        & & \texttt{rag} & 65.2 & 64.2 & 504 & 506 & 4 \\
        \\
        %\multirow{9}{*}{Grid}
        %& \multirow{3}{*}{Linear}
        %  & \texttt{fcg} & - & - & - & - & - \\
        %& & \texttt{knn} & - & - & - & - & - \\
        %& & \texttt{rag} & - & - & - & - & - \\
        \multirow{6}{*}{Grid}
        & \multirow{3}{*}{SinCos}
          & \texttt{fcg} & 70.8 & 69.4 & 388 & 418 & 10 \\
        & & \texttt{knn} & 72.2 & 71.8 & 450 & 475 & 12 \\
        & & \texttt{rag} & \sndbest{74.3} & 73.1 & 272 & 420 & 8 \\
        & \multirow{3}{*}{BERT}
          & \texttt{fcg} & $\dagger70.6$ & 69.5 & 479 & 512 & 6 \\
        & & \texttt{knn} & 71.7 & 69.5 & 253 & 260 & 6 \\
        & & \texttt{rag} & *\fstbest{75.5} & 72.2 & 492 & 512 & 8 \\
    \bottomrule
    \end{tabular}
    \caption{Ablation Study for each superpixel generation strategy $S$, each positional embedding $P$, and each graph connectivity strategy $G$ for the FashionMNIST and CIFAR10 datasets. We mark the \fstbest{best} and \sndbest{second best} validation accuracies for each dataset. The best SLIC-based model and non-ViT grid-based is marked with *, and the original ViT model is marked with $\dagger$. For the CIFAR10 dataset Linear embeddings were failing and thus were cancelled to save compute time.}
    \label{tab:ablation}
\end{table*}

We trained our SPT-B model with patch size of $S\approx10$ for about 150 on the Resisc45 dataset using a NVIDIA GTX 1070m, after which it achieved a validation accuracy of 53.5\%, which only slightly lower performance to that reported for the much larger ViT-B with a patch size of 16 trained from scratch for a similar amount of hours on a higher-end GPU. When testing our model on the Imagenette dataset a SLIC-based Ti model with Sine-Cosine positional embedding achieved 67.7\% test accuracy while a grid-based Sine-Cosine model achieved 67.3\%.

\begin{figure*}
    \centering
    \includegraphics[width=0.45\linewidth]{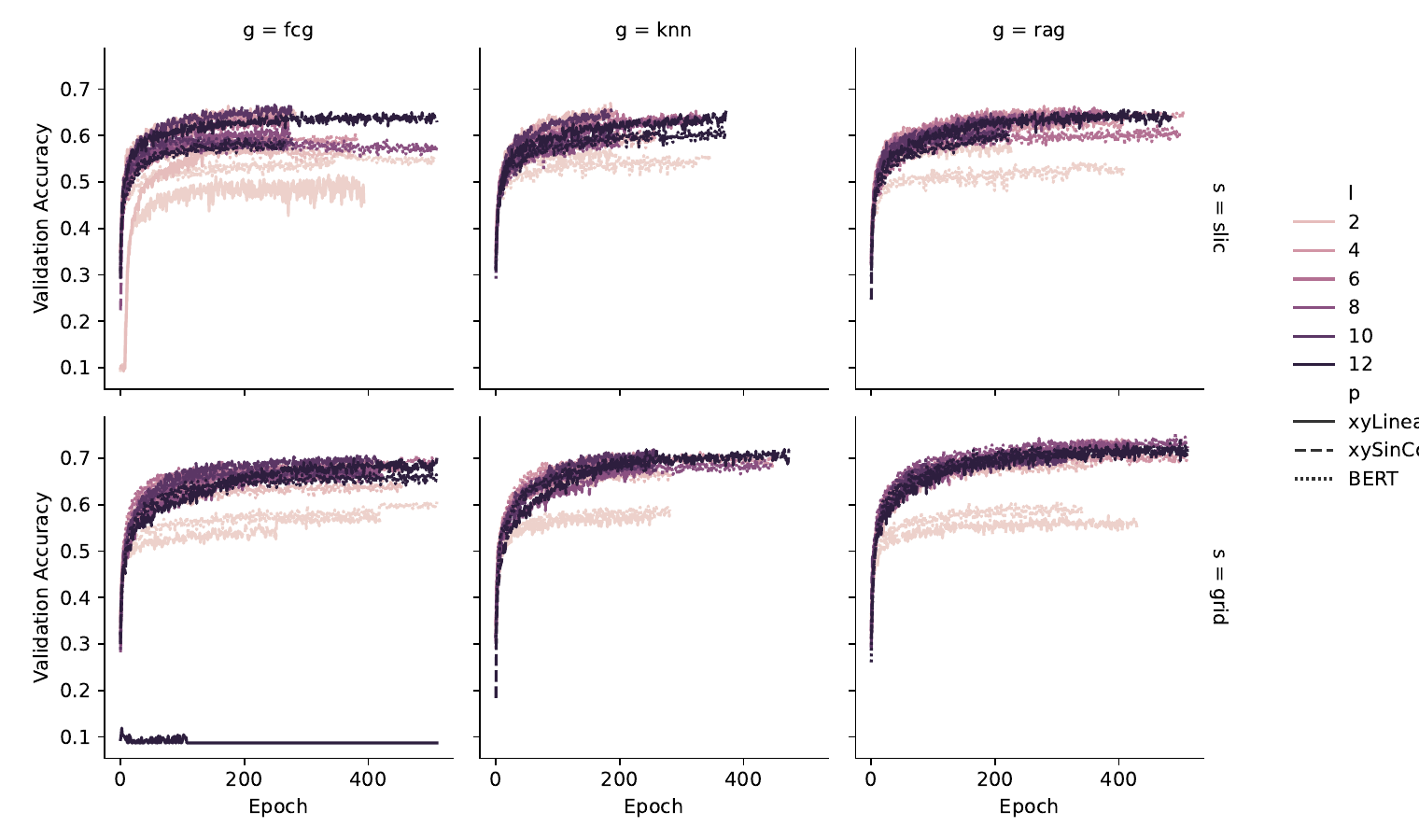}
    \includegraphics[width=0.45\linewidth]{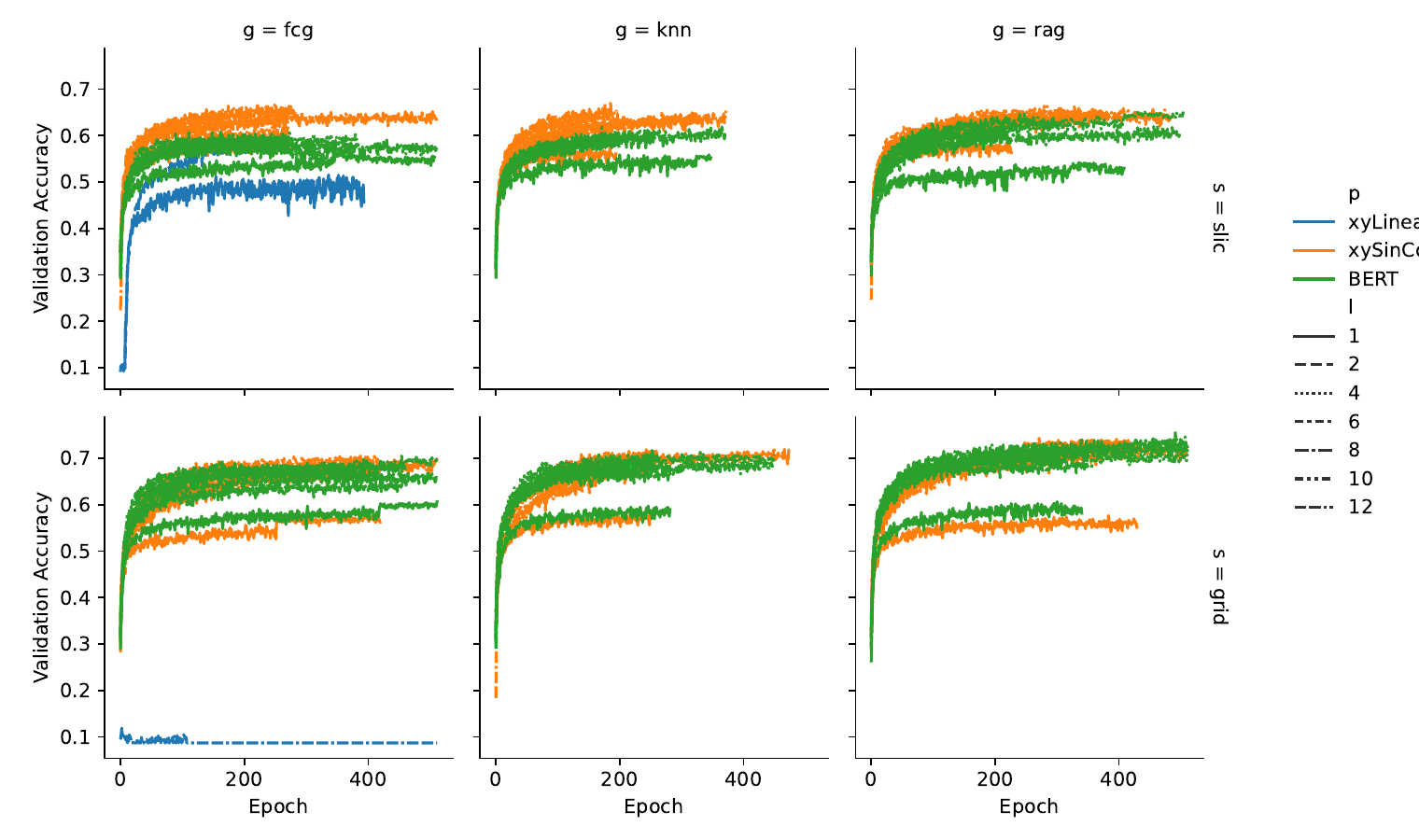}
    \caption{Plots highlighting the Validation Accuracy throughout the epochs for all the configurations we trained on the CIFAR10 dataset. One can see from graph on the left that the number of layers seemed to be more relevant for this dataset than the FashionMNIST dataset, possibly due to this dataset requiring a higher model capacity. Higher numbers of layers, however, often had a diminishing returns effect and sometime even negatively affected performance. For the slic model, the BERT encoding often fared worse than the Sine-Cosine embedding based on the xy coordinates. This is likely due to the fact that the superpixel embeddings have variable xy coordinates and thus the fixed learned positional encoding cannot work properly. Due to resource-sharing on HPC many models ran for fewer epochs, an analysis of time complexity and time per iteration can be see in Table~\ref{tab:fashionmnist-layer-time} and Figure~\ref{fig:fashionmnist-layer-time} where we analyse based on the FashionMNIST dataset, which ran on a more controlled environment. The linear xy positional embedding exhibited inconsistent performance and was dropped from further experiments early on.}
    \label{fig:cifar10-ablation}
\end{figure*}

\section{Conclusion}

In this paper we proposed a generalisation of the Vision Transformer model to accept arbitrary superpixel formats and graph connectivities, bridging the gap between using Vision Transformers and the Superpixel-based Image Classification literature. We've tested our models with two different Superpixel generation strategies, three graph connectivity strategies, and three positional embedding strategies, showing through this ablation study that our proposed multidimensional SinCos worked better for both, and that graph connectivity strategies produce different results in different datasets.

Furthermore, we validated our model against three previous works on Superpixel Image Classification and show better performance than these baselines, showing that even when constrained to working under a superpixel framework (e.g. if one wants to transfer between euclidean and non-euclidean images as was proposed in \cite{avelar_superpixel_2020}, one could use a similar approach to ours and would only need to adapt the initial embedding head, keeping the transformer core of the model. We test our model on datasets with different image sizes and sample contents, showing that our model can adapt to these different circumstances.

Our model, however, does not come without compromises. To be able to fully utilise the superpixel information an extra channel was needed during the patch embedding and the lack of search distance limitations on the SLIC implementation used meant that our patches could have sizes above the $2S$ limit defined in the original SLIC/SLIC0 algorithm \cite{achanta_slic_2010}. Furthermore, since the number of superpixels is dynamic and might be different than the value specified, it was necessary to provide padding on the input sequences. These issues, however, did not constitute a major resource consumption increase in our model. Our experiments were also heavily limited due to the compute available to us, which we hope to overcome on future work with access to better compute.

Future work can address some of the abovementioned weaknesses of our models and can also attempt to use convolutional cores in the patch embedding heads \cite{xiao_early_2021}, as these would allow the model to better identify the superpixel's component parts and take full advantage of the superpixel's shape and colour information. Finally, future work could also focus on pre-training a Superpixel-based model, since pre-training is known to be a much better strategy for training transformer-based image models \cite{steiner_how_2022}, this would also allow one to attempt transferring models between euclidean and non-euclidean domains as mentioned above, which our model is naturally more suitable for due to its flexible learning pattern, even though our results are only comparable to a standard ViT model.

\sndefinitions

%===========================================================================================%%

\bibliographystyle{plain}
\bibliography{gat2vit}

\clearpage
\newpage

\begin{appendices}
\onecolumn

\setcounter{table}{0}
\renewcommand{\thetable}{A\arabic{table}}
\setcounter{figure}{0}
\renewcommand{\thefigure}{A\arabic{figure}}
\setcounter{algorithm}{0}
\renewcommand{\thealgorithm}{A\arabic{algorithm}}

\section{}

\begin{algorithm*}
\label{alg:squaregrid}
\caption{Square Grid Superpixel}
\begin{algorithmic}[1]
\Require $\text{image} \in \mathbb{R}^{C \times h \times w}$, $\text{patch\_size} \in \mathbb{Z}$
\Ensure $\text{segments} \in \mathbb{Z}^{h \times w}$
\Ensure $\sqrt{\text{n\_segments}} \in \mathbb{Z}$
\Ensure $H \mod \text{patch\_size} = 0$
\Ensure $H = W$
\State $\text{num\_patches\_per\_line} \gets \sqrt{\text{n\_segments}}$
\State Initialize $\text{segments} \gets \text{zeros}(H, W)$
\For{$i \gets 0$ to $\text{n\_segments} - 1$}
    \State $i_{y} \gets \lfloor i/\text{num\_patches\_per\_line} \rfloor$
    \State $i_{x} \gets i \mod \text{num\_patches\_per\_line}$
    \State Assign segment index $i$ to region in $\text{segments}$
    \State $\text{segments}[i_{y}\times\text{patch\_size} \ldots (i_{y}+1)\times\text{patch\_size}, i_{x}\times\text{patch\_size} \ldots (i_{x}+1)\times\text{patch\_size}] \gets i$
\EndFor
\State \Return $\text{segments}$
\end{algorithmic}
\end{algorithm*}

\begin{algorithm*}
\label{alg:xysincos}
\caption{Sine-Cosine Coordinate Encoding}
\begin{algorithmic}[1]
\Require $\text{hidden\_dim} \in \mathbb{Z}$, $\text{n\_axes} \in \mathbb{Z}$
\Ensure $\text{pos\_emb} \in \mathbb{R}^{b \times s \times h}$, where $b$ is batch size, $s$ is sequence length, $h$ is hidden dimension
\State Initialize $\text{div\_term} = \exp([0 \ldots h] \cdot (-\log(10000) / h))$
\State Initialize $\text{mixer} = W \in \mathbb{R}^{h\times h}$
\Procedure{Encode}{$\text{pos}, \text{x}$}
    \Comment{\text{pos} is $\mathbb{R}^{b \times s \times d}$, \text{x} is $\mathbb{R}^{b \times s \times h}$}
    \State $d \gets \min(\text{n\_axes}, \text{pos.shape}[2])$
    \State Initialize $\text{pos\_emb} \gets \text{zeros\_like}(\text{x})$
    \For{$i \gets 0$ to $d - 1$} \Comment{Build sin-cos for each dimension}
        \State Select all indices for cosine as $j_{\text{cos}} = \{i, i + 2d, i + 4d, \ldots\}$ up to $h$
        \State Select all indices for sine as $j_{\text{sin}} = \{i + 1, i + 1 + 2d, i + 1 + 4d, \ldots\}$ up to $h$
        \State $\text{pos\_emb}[:,:,j_{\text{cos}}] \gets \cos(\text{pos}[:,:,i] \cdot \text{div\_term}[j_{\text{cos}}])$
        \State $\text{pos\_emb}[:,:,j_{\text{sin}}] \gets \sin(\text{pos}[:,:,i] \cdot \text{div\_term}[j_{\text{sin}}])$
    \EndFor
    \State \Return $\text{mixer}(\text{pos\_emb})$
\EndProcedure
\end{algorithmic}
\end{algorithm*}

\begin{table*}[]
    \footnotesize
    \centering
        \begin{tabular}{lllrr}
    \toprule
    S & P & G & $\mu(b)$ & $\sigma(b)$ \\
    \midrule
    \multirow[t]{9}{*}{slic} & xyLinear & fcg & 2191.445 & 79.985 \\
    
     & xySinCos & fcg & 2163.883 & 56.462 \\
    
     & BERT & fcg & 2163.673 & 91.721 \\
    
     & xyLinear & rag & 3018.270 & 67.750 \\
    
     & xySinCos & rag & 2890.867 & 59.136 \\
    
     & BERT & rag & 2945.360 & 97.905 \\
    
     & xyLinear & knn & 3665.725 & 59.916 \\
    
     & xySinCos & knn & 3784.874 & 75.899 \\
    
     & BERT & knn & 3622.777 & 38.381 \\
    
    \multirow[t]{9}{*}{grid} & xyLinear & fcg & 2308.859 & 53.799 \\
    
     & xySinCos & fcg & 2198.941 & 40.184 \\
    
     & BERT & fcg & 2127.548 & 68.252 \\
    
     & xyLinear & rag & 2927.251 & 51.175 \\
    
     & xySinCos & rag & 2820.486 & 32.625 \\
    
     & BERT & rag & 2860.717 & 70.748 \\
    
     & xyLinear & knn & 3547.717 & 97.029 \\
    
     & xySinCos & knn & 3590.628 & 77.070 \\
    
     & BERT & knn & 3636.872 & 133.942 \\
    
    \multirow[t]{3}{*}{slic} & xyLinear & * & 2958.480 & 252.643 \\
    
     & xySinCos & * & 2946.541 & 274.293 \\
    
     & BERT & * & 2910.603 & 244.672 \\
    
    \multirow[t]{3}{*}{grid} & xyLinear & * & 2927.942 & 251.979 \\
    
     & xySinCos & * & 2870.018 & 252.923 \\
    
     & BERT & * & 2875.046 & 261.837 \\
    
    \multirow[t]{3}{*}{slic} & \multirow[t]{3}{*}{*} & fcg & 2173.000 & 40.673 \\
     &  & rag & 2951.499 & 42.239 \\
     &  & knn & 3691.125 & 43.873 \\
    
    \multirow[t]{3}{*}{grid} & \multirow[t]{3}{*}{*} & fcg & 2211.783 & 35.102 \\
     &  & rag & 2869.485 & 30.399 \\
     &  & knn & 3591.739 & 56.513 \\
    
    \multirow[t]{9}{*}{*} & \multirow[t]{3}{*}{fcg} & xyLinear & 2250.152 & 52.541 \\
     &  & xySinCos & 2181.412 & 40.801 \\
     &  & BERT & 2145.611 & 54.348 \\
    
     & \multirow[t]{3}{*}{rag} & xyLinear & 2972.761 & 45.539 \\
     &  & xySinCos & 2855.677 & 40.709 \\
     &  & BERT & 2903.039 & 66.223 \\
    
     & \multirow[t]{3}{*}{knn} & xyLinear & 3606.721 & 55.538 \\
     &  & xySinCos & 3687.751 & 71.164 \\
     &  & BERT & 3629.824 & 65.661 \\
    
    slic & * & * & 2938.542 & 144.431 \\
    
    grid & * & * & 2891.002 & 143.395 \\
    
    \multirow[t]{6}{*}{*} & xyLinear & * & 2943.211 & 174.740 \\
    
     & xySinCos & * & 2908.280 & 183.329 \\
    
     & BERT & * & 2892.825 & 175.366 \\
    
     & \multirow[t]{3}{*}{*} & fcg & 2192.391 & 28.569 \\
     &  & rag & 2910.492 & 29.785 \\
     &  & knn & 3641.432 & 36.430 \\
    
    \bottomrule
    \end{tabular}
    \caption{Table showing the average and standard deviation of the intercepts for linear regression models build on a certain combination of Superpixel, Positional Embedding and Graph Adjacency method. We can see that the Graph Adjacency method has the biggest impact on performance. Values for slopes are not included since most slopes had a non-significant p-value, and the comparison between intercepts yielded more significant p-values.}
    \label{tab:fashionmnist-layer-time}
\end{table*}

\begin{figure*}
    \centering
    \includegraphics[width=0.5\linewidth]{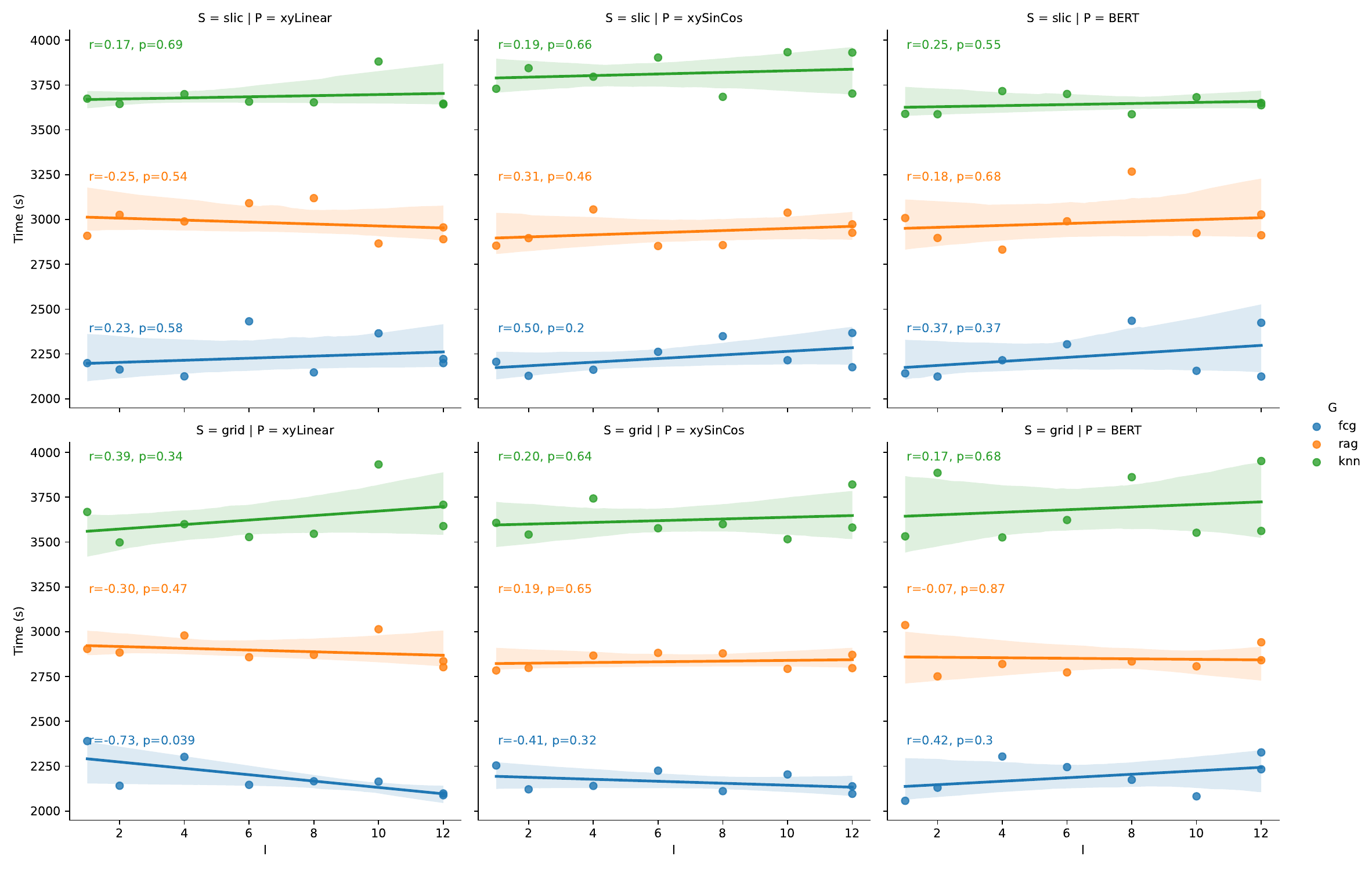} \\
    \begin{tabular}{ccc}
    \includegraphics[width=0.3\linewidth]{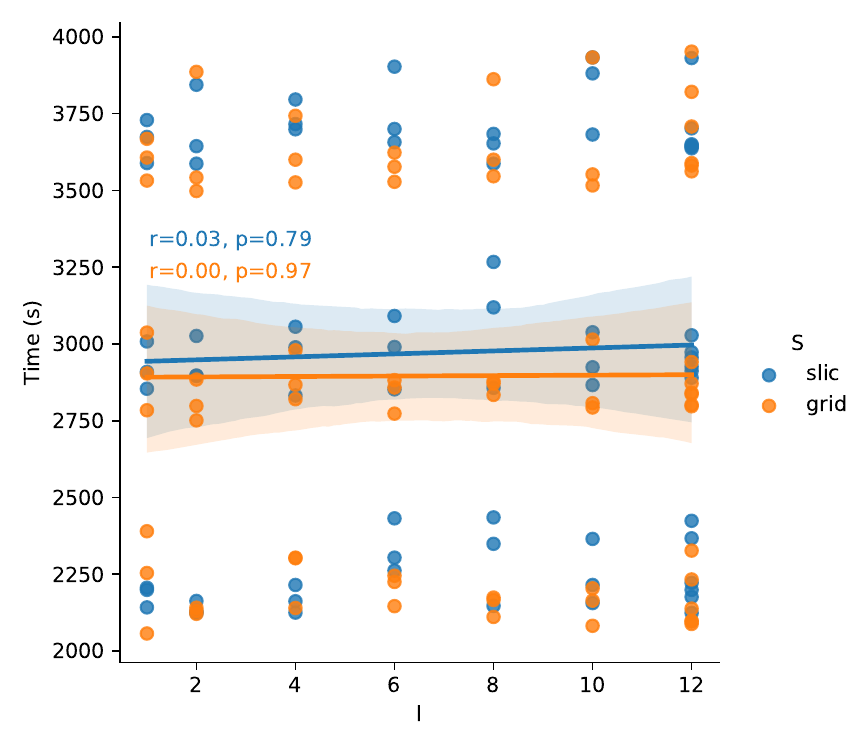}
    &
    \includegraphics[width=0.3\linewidth]{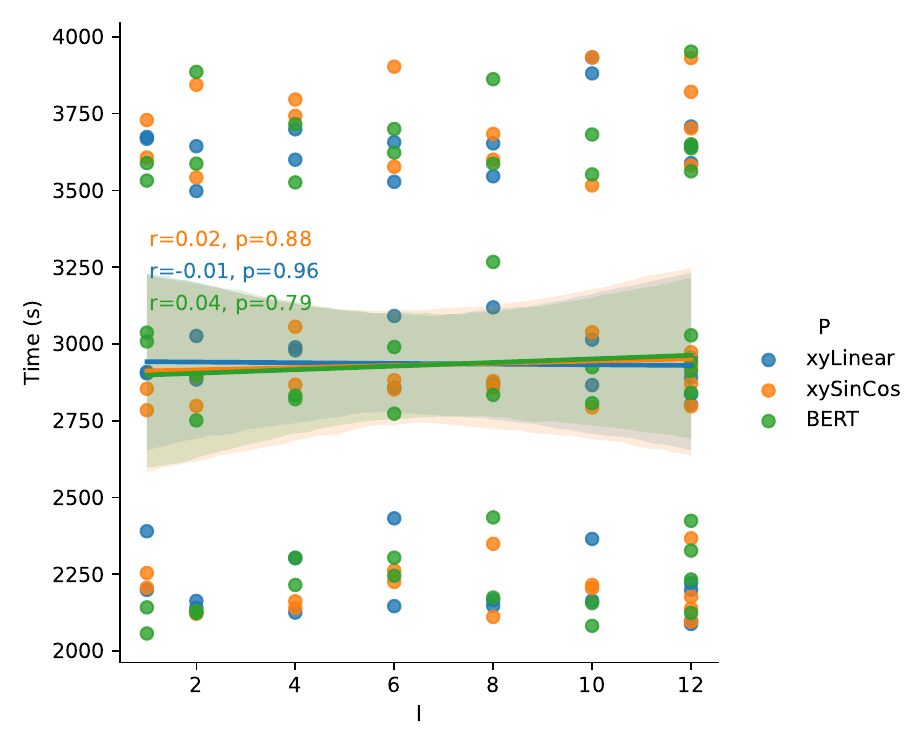}
    &
    \includegraphics[width=0.3\linewidth]{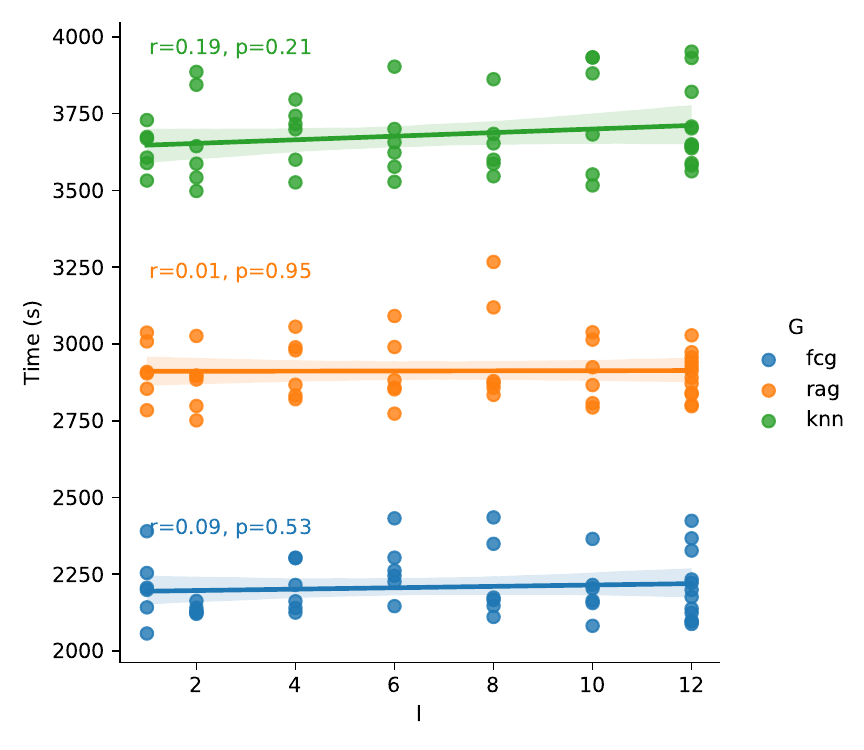}
    \end{tabular}
    \caption{Figures showing regression plots (with person r and p values) between the number of layers and the time taken for the process to finish training and testing. The top figure shows results separated by each of the variables, and most results show a positive r but a not-significant p, with the only significant p having a negative r, which we believe to be spurious. Only when we look at each variable individually on the bottom figures that we are able to find significant differences between variables, and the graph building algorithm is has the biggest difference.}
    \label{fig:fashionmnist-layer-time}
\end{figure*}
\end{appendices}

\end{document}